\pdfoutput=1
\documentclass[11pt]{article}
\PassOptionsToPackage{hyperfootnotes=false}{hyperref}
\makeatletter
\AtBeginDocument{%
  \@ifundefined{hyper@natlinkstart}{}{%
    \def\hyper@natlinkstart#1{}\def\hyper@natlinkend{}%
    \def\hyper@natlinkbreak#1#2{#1}}}%
\makeatother

\IfFileExists{acl.sty}{%
  \usepackage[preprint]{acl}  
  
}{%
  \usepackage[a4paper,margin=2.3cm]{geometry}
  \usepackage[numbers,sort&compress]{natbib}
  \setcitestyle{authoryear,round}
  \usepackage{multicol}
  
}

\usepackage{times}
\usepackage{latexsym}
\usepackage[T1]{fontenc}
\usepackage[utf8]{inputenc}
\usepackage{microtype}
\IfFileExists{inconsolata.sty}{\usepackage{inconsolata}}{}
\usepackage{amsmath,amssymb,bm}
\usepackage{booktabs}
\usepackage{multirow}
\usepackage{graphicx}
\usepackage{xcolor}
\usepackage{enumitem}

\newcommand{\modl}{\textsc{Modl}}
\newcommand{\fic}{\textsc{Fic}}
\newcommand{\dlora}{\mbox{D-LoRA}}
\newcommand{\llora}{\mbox{L-LoRA}}

\title{One Domain, Many Tongues: Composing Domain and Language LoRAs \\
for Cross-Lingual Remote-Sensing MLLMs without Paired Data}

\author{Xuechen Li \\
  University of Minnesota, Twin Cities \\
  \texttt{li003487@umn.edu}}

\begin{document}
\maketitle

\begin{abstract}
Remote-sensing (RS) multimodal large language models (MLLMs) are
trained and evaluated only in
English, while text-only instruction data covers over 100 languages.
We propose \modl{} (\emph{M}utually \emph{O}rthogonal
\emph{D}omain--\emph{L}anguage composition), a recipe that adds new
languages to an English RS MLLM without a single multilingual RS
example: a domain LoRA trained on English RS imagery and a language
LoRA trained on text alone are learned jointly, under one loss term
that keeps the two updates mutually orthogonal at every layer
throughout training. This constraint is the recipe's active
ingredient. Without it, the same training answers RS questions
correctly \emph{but in English}, erases much of the base model's
multilingual text ability, and diverges on one seed in three; sixteen
alternatives, from training-free merging to prior orthogonality
variants, fail the same way. \modl{}
repairs every failure on every seed: answers are correct and in the
target language 56--71\% of the time, where the best alternative
reaches 27\% and most stay below 8\%, text ability stays at the level of the untrained base,
and on Spanish it surpasses Qwen2.5-VL-7B, with zero
multilingual--multimodal data. A single five-language adapter
retains English, Spanish, and Vietnamese at full strength across
three seeds; non-Latin scripts remain an open boundary.\footnote{Code, adapters, and the preregistered evaluation
protocol will be released upon publication.}
\end{abstract}

\section{Introduction}
\label{sec:intro}

\begin{figure}[t]
\centering
\includegraphics[width=\columnwidth]{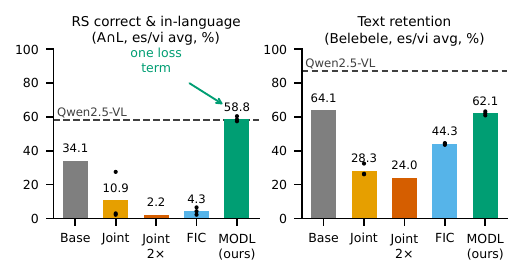}
\caption{\textbf{One loss term changes everything.} \textbf{Left}:
correct-and-in-language accuracy (A$\cap$L) on the GeoChat-Bench
classification holdout (2{,}195 items per language, es/vi average).
\textbf{Right}: text-only multilingual competence, measured as
likelihood-ranked Belebele accuracy (900 items per language, es/vi
average), which is immune to generation-side defects. Bars are means
over three seeds; black dots mark individual seeds; the dashed line
is the Qwen2.5-VL-7B reference. One mutual-orthogonality loss term
(\modl{}, green) separates the only working system from every
alternative. The high \textsc{Joint} seed dot on the left is the
diverged seed's step-500 substitute (App.~\ref{app:full}).}
\label{fig:teaser}
\end{figure}

Remote-sensing (RS) multimodal large language models (MLLMs) are
built and evaluated in English
\citep{kuckreja2024geochat, muhtar2024lhrs, zhang2024earthgpt,
soni2025earthdial}. The standard route to multilinguality trains on
millions of machine-translated (MT) multimodal instructions
\citep{maaz2025palo, yue2025pangea, geigle2025centurio}; transplanted
to a vertical domain, its cost multiplies by the number of
languages.\footnote{The only partly non-English RS resources are
English--Chinese \citep{zhou2025brsic, wang2025xlrsbench}; no RS
resource covers a low-resource language.} The modular parameter-efficient fine-tuning (PEFT) literature
promises a far cheaper route: train a \emph{domain} LoRA (low-rank
adaptation; \citealp{hu2022lora}) on English RS
instructions and a \emph{language} LoRA on text-only instructions
\citep{li2023bactrianx}, and compose them \citep{pfeiffer2020madx,
ansell2022composable, chronopoulou2024language, bandarkar2025layer}.
That literature, however, is text-only, and two obstacles stand
between it and the multimodal setting: LLaVA-style models drift to
English the moment an image enters the context (image-induced
fidelity loss, IFL; \citealp{hinck2024ifl}),
and independently trained LoRA updates interfere when summed
\citep{stoica2025knots}.

We propose \modl{}: joint training of the two adapters under a
symmetric \emph{mutual-orthogonality} constraint, a single loss term
that keeps the domain and language updates in disjoint subspaces at
every layer throughout training. We validate \modl{} in the first
systematic, preregistered study of domain$\times$language LoRA
composition in an MLLM: LLaVA-1.5-7B, five typologically diverse
languages, a matched training budget for every system, three seeds,
and decision rules frozen in advance. The comparison spans every family of
alternatives: plain joint training, fourteen training-free mergers,
prior orthogonal-training methods (OSRM, O-LoRA), and \fic{}, a
function-space consistency variant representing the behavioral
alternative suggested by recent skepticism that geometry is what
controls interference \citep{zhang2025rethinking}.

The comparison is one-sided (Fig.~\ref{fig:teaser}). Capability
composes but language does not: naively co-trained adapters answer
Spanish and Vietnamese RS questions correctly \emph{in English}, a
fidelity failure that scoring against translated references mistakes
for incapacity. Naive joint training is also destructive: it erases
the base model's text-only multilingual competence, degrades even its
native target-language RS accuracy, and diverges on one of three
seeds. Among the alternatives, \fic{} repairs only the
text-competence loss. \modl{}, one loss term, repairs everything, on
every seed and across two orders of magnitude of its single
hyperparameter; on Spanish scene classification it surpasses even
Qwen2.5-VL-7B's incidental multilinguality in
correct-and-in-language accuracy (A$\cap$L).

\medskip
Contributions:
\begin{enumerate}[itemsep=1pt,topsep=2pt,leftmargin=*]
\item \textbf{The first cross-lingual RS MLLM recipe}: mutually
  orthogonal joint training of a domain and a language LoRA, the only
  trained system that beats the untrained base on A$\cap$L
  (\S\ref{sec:results}).
\item \textbf{A diagnosis that reframes the problem}: the obstacle
  to composition is interference and output-language fidelity, not
  capability transfer (\S\ref{sec:results}).
\item \textbf{Mechanism evidence}: under one matched budget, a
  function-space variant with the same goal (\fic{}) and prior
  initialization-time and sequential orthogonality (OSRM, O-LoRA) all
  fail. What matters is that the constraint is mutual and active
  throughout training, contrary to text-only evidence on
  orthogonality \citep{zhang2025rethinking}
  (\S\ref{sec:method:fic}).
\item \textbf{A preregistered protocol with two quantified evaluation
  pitfalls} (missing end-of-sequence (EOS) supervision, $7\times$ metric corruption;
  inconsistently translated references, up to 41\% of items unscorable) and
  corrected class-identity scoring (\S\ref{sec:setup}).
\end{enumerate}

\section{Related Work}
\label{sec:related}

\paragraph{Remote-sensing MLLMs.}
GeoChat \citep{kuckreja2024geochat} established LoRA-tuning of a
LLaVA-family model \citep{liu2023llava} on RS instructions, followed by
RSGPT \citep{hu2023rsgpt}, LHRS-Bot \citep{muhtar2024lhrs}, EarthGPT
\citep{zhang2024earthgpt}, EarthDial \citep{soni2025earthdial}, and
many successors. All are English-only.
Multilingual RS resources are embryonic: EN--ZH captions and
annotations (BRSIC, XLRS-Bench; \citealp{zhou2025brsic,
wang2025xlrsbench}), the RS-M-CLIP encoder \citep{silva2024rsmclip},
retrieval-based training-free captioning
\citep{rebelo2025multilingual}, and translation-as-augmentation for
RSVQA \citep{yuan2023multilingual}. None yields an
instruction-following RS assistant in any non-English language.

\paragraph{Multilingual MLLMs and IFL.}
PALO \citep{maaz2025palo}, mBLIP \citep{geigle2024mblip}, Pangea
\citep{yue2025pangea}, and Centurio \citep{geigle2025centurio} train
on multilingual multimodal data; VisCPM \citep{hu2024viscpm} exploits a
multilingual backbone; \citet{pikabea2025breaking} inject text-only
multilingual data \emph{during} visual supervised fine-tuning (SFT)
to mitigate IFL
\citep{hinck2024ifl}; Aya Vision \citep{dash2025ayavision} merges to
\emph{preserve} pre-existing multilinguality. None adds a new language
to an English-centric MLLM by composing a text-only language module,
and none addresses a specialized visual domain.

\paragraph{Modular composition for cross-lingual transfer.}
MAD-X \citep{pfeiffer2020madx} and LT-SFT \citep{ansell2022composable}
compose language and task modules in encoder models; the LLM-era line
includes language/task arithmetic \citep{chronopoulou2024language},
AdaMergeX \citep{zhao2025adamergex}, expert layer swapping
\citep{bandarkar2025layer}, and language--task LoRA interpolation
\citep{lee2025mlm}. xGQA \citep{pfeiffer2022xgqa}
stacked adapters for cross-lingual visual question answering (VQA)
in pre-LLM models. All of this
is text-only or pre-LLM; the multimodal vertical-domain instantiation
is our subject.

\paragraph{Merging, merge-aware training, and function-space fusion.}
Training-free mergers reduce interference at merge time: task
arithmetic \citep{ilharco2023task}, TIES \citep{yadav2023ties}, DARE
\citep{yu2024dare}, AdaMerging's layer-wise coefficients
\citep{yang2024adamerging}, KnOTS' SVD alignment
\citep{stoica2025knots}, and CAT's learned concatenation
\citep{prabhakar2024cat}; surveys in \citet{yang2024mergesurvey}. In
MLLMs, merging composes heterogeneous models \citep{du2025adamms} and
transfers multilinguality from a multilingual LLM's residuals
\citep{wang2026dim3}.
Merge-\emph{aware training} makes models composable by construction:
fixed orthogonal bases (orthogonal adaptation, QR-LoRA;
\citealp{po2024orthogonal, yang2025qrlora, dyme2025}), data-aware pre-fine-tuning constraints \citep{zhang2025osrm},
sharpness-aware fine-tuning \citep{lee2025saft}, linearized
fine-tuning \citep{ortiz2023tangent}, and sequential inter-LoRA
orthogonality for continual learning \citep{wang2023olora}; all are
parameter-geometric. \citet{zhang2025rethinking} report that strict
inter-LoRA orthogonality alone does not deliver compositionality in
text-only merging; our multimodal evidence points the other way, and
sharpens \emph{which} orthogonality matters: we compare
initialization-time (OSRM), sequential (O-LoRA), and our
\emph{mutual, simultaneous} constraint under one budget, and only the
last restores language fidelity (Table~\ref{tab:main}); the prior
families reproduce the naive-joint failure signature.
\fic{} instead belongs to the \emph{function-space} family:
RegMean matches linear-layer outputs at merge time \citep{jin2023regmean},
FuseLLM distills several source LLMs into one via continual training
\citep{wan2024fusellm}, and multi-teacher distillation has been used to
repair merged models post hoc \citep{mtkd2025}; Learning-without-
Forgetting-style consistency preserves old behavior in continual
learning \citep{li2017lwf}.
\fic{} differs from all of these in that the consistency is imposed
\emph{during} adapter training, on the model's \emph{own} unpaired
training streams, with the detached single-adapter configuration as
teacher (no transfer set, no separate distillation phase, no external
teacher), and in that the composed axes are domain and language in a
multimodal LLM.

\section{Method}
\label{sec:method}

\subsection{Problem Setup}
\label{sec:method:setup}

Let $f_{W_0}$ be an English-centric MLLM (LLaVA-1.5-7B
\citep{liu2023llava}: CLIP ViT-L/336, projector, Vicuna-7B) with
parameters $W_0$. The English-centric backbone is a deliberate
control: it isolates the language-acquisition effect that a strongly
multilingual backbone would confound (we include such a backbone as a
reference point, \S\ref{sec:setup}). We are given (i) an English RS
instruction corpus $\mathcal{D}_{\mathrm{dom}}$
(image--instruction--response triples) and (ii) per target language
$\ell \in \mathcal{L}$ a \emph{text-only} instruction corpus
$\mathcal{D}_{\ell}$ with no RS content and no images. The goal is a
single model that answers RS questions posed in $\ell$, \emph{in}
$\ell$, having never seen an RS example in $\ell$ or any
$\ell$-language image--text pair, with no test-time routing, language
detection, or per-input adapter switching.

\subsection{Domain and Language LoRAs}
\label{sec:method:loras}

Each adapted weight $W^{(m)}_0 \in \mathbb{R}^{d \times k}$ receives
LoRA \citep{hu2022lora} updates
$\Delta W^{(m)} = \tfrac{\alpha_r}{r} B^{(m)} A^{(m)}$.
The \dlora{} $\{\Delta W_D^{(m)}\}$ trains on
$\mathcal{D}_{\mathrm{dom}}$ and attaches to attention and MLP
projections of all LLM layers and to the vision--language projector.
Each \llora{} $\{\Delta W_{L_\ell}^{(m)}\}$ trains on
$\mathcal{D}_{\ell}$ (no images) and attaches to the LLM only,
because a text-only corpus provides no signal for visual alignment.
At inference the two updates are simply added into the base weights
(\S\ref{sec:method:merge}).

\subsection{Joint Training and the Geometric Hypothesis}
\label{sec:method:geo}

Training alternates micro-batches from $\mathcal{D}_{\mathrm{dom}}$
(updating $\theta_D$) and from the language corpora (updating
$\theta_L$); within a run the target languages share a single
\llora{} trained on their union (es+vi in the core setting, one
language in per-pair runs, five in the five-way variant), so
\dlora{} is optimized to coexist with the language subspace it will
be deployed with. We
call this plain scheme \textsc{Joint}. It already differs from
independent training (the input to all merging baselines) by letting
the optimizer see both objectives, though never a paired example.

\begin{figure*}[t]\centering
\includegraphics[width=\textwidth]{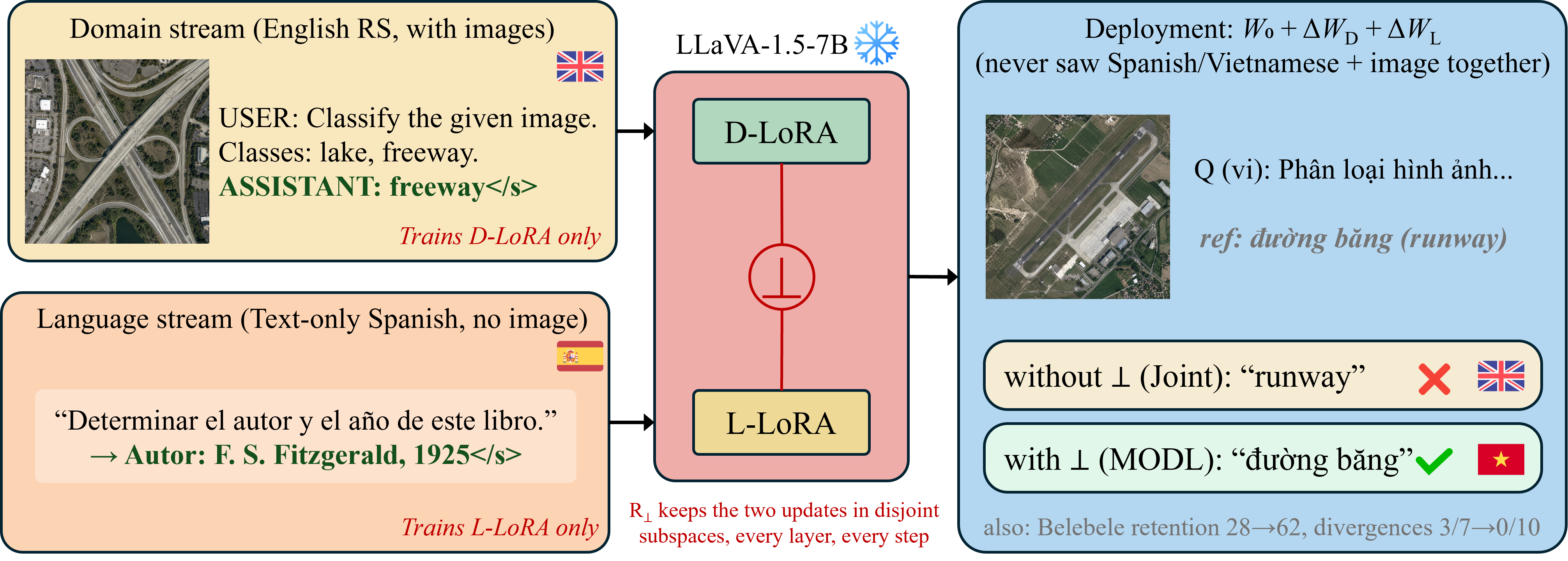}
\caption{\textbf{The \modl{} recipe, shown on real training and test
examples.} \textbf{Left}: the two unpaired training streams as they
actually appear to the model: an English RS instruction with its
image (supervising \dlora{} only) and a text-only Spanish
instruction with no image (supervising \llora{} only); targets end
with the supervised EOS. \textbf{Center}: both adapters are trained
jointly on the frozen backbone; the mutual orthogonality penalty
$\mathcal{R}_\perp$ (Eq.~\ref{eq:orth}) keeps the two updates in
disjoint subspaces at every layer and every step. \textbf{Right}:
deployment simply adds both updates,
$W_0+\Delta W_{\mathrm{D}}+\Delta W_{\mathrm{L}}$, and is queried on
a combination never seen in training --- a Vietnamese question about
an RS image. Without the constraint (\textsc{Joint}) the model names
the correct class in English; with it (\modl{}) it answers correctly
in Vietnamese. The footer notes the other two repairs: multilingual
text retention (Belebele) and optimization stability (no constrained
run diverged, vs.\ 3 of 7 for the naive recipe).}
\label{fig:pipeline}
\end{figure*}

The merge-aware-training literature predicts that interference should
be removed by \emph{decoupling parameter subspaces}. Prior
instantiations constrain sequential task streams \citep{wang2023olora},
initialization-time subspaces \citep{zhang2025osrm}, or fixed bases
\citep{po2024orthogonal}; none address two heterogeneous modules
trained simultaneously. Our instantiation, \modl{}, adds a
symmetric \emph{mutual} penalty on row/column-normalized factors of
the two adapters at every layer,
\begin{equation}
\label{eq:orth}
\mathcal{R}_{\perp}^{(m)} =
\big\| \hat{A}_D^{(m)} \hat{A}_{L_\ell}^{(m)\top} \big\|_F^2
+ \big\| \hat{B}_D^{(m)\top} \hat{B}_{L_\ell}^{(m)} \big\|_F^2,
\end{equation}
driving $\Delta W_D^{\top}\Delta W_L \!\approx\! 0$ and
$\Delta W_D \Delta W_L^{\top} \!\approx\! 0$, weighted by
$\lambda_{\perp}$ (0.1 primary; results are stable across
$\lambda_{\perp}\!\in\![0.01,1]$) and trained identically to
\textsc{Joint} otherwise. \modl{} is the paper's
proposed recipe; Figure~\ref{fig:pipeline} sketches the training
setup and the three failure modes the constraint removes. To test whether its effect is genuinely geometric,
we also construct the natural function-space alternative, \fic{}
below, which pursues the same non-interference goal without
constraining where the modules live; the contrast between the two
carries the mechanistic claim. (The preregistration originally
favored the functional hypothesis; the reversal is documented in
App.~\ref{app:hyper}.)

\subsection{\fic{}: Functional Interference Consistency}
\label{sec:method:fic}

What we actually require of composition is behavioral, not geometric:
\emph{on domain inputs, adding the language module must not change what
the domain module does; on language inputs, vice versa}. \fic{}
imposes exactly this, on the two unpaired streams already present in
joint training.

For a micro-batch from $\mathcal{D}_{\mathrm{dom}}$, let the
\emph{teacher} be the model with only the domain adapter active,
$p_T = f_{W_0 + \Delta W_D}$, with gradients stopped, and the
\emph{student} be the additive composition
$p_S = f_{W_0 + \Delta W_D + \Delta W_{L_\ell}}$. With $S$ the set of
supervised (response) token positions, capped at $K$ per micro-step,
\begin{equation}
\label{eq:fic}
\mathcal{L}_{\fic}^{\mathrm{dom}} = \tfrac{1}{|S|} \textstyle\sum_{t \in S}
\mathrm{KL}\!\left( p_T(\cdot | x, y_{<t}) \,\|\,
                    p_S(\cdot | x, y_{<t}) \right),
\end{equation}
at temperature $\tau$. Symmetrically, on a batch from
$\mathcal{D}_{\ell}$ the teacher is the language-only model
$f_{W_0 + \Delta W_{L_\ell}}$ and the student is the same composition.
The total objective per stream is the task cross-entropy of that
stream's own adapter plus
$\lambda_{\fic}\, \mathcal{L}_{\fic}$.
Frozen configuration from the preregistered pilot:
$\lambda_{\fic}{=}0.5$, $\tau{=}1.0$, $K{=}64$, applied from the
first step with no warm-up or ramp.

Three properties matter. \emph{(i) No pairing:} the two consistency
terms each use only their own stream; no multilingual RS example, real
or synthetic, is ever constructed. \emph{(ii) No routing:} the
deployed model is the plain additive composition; \fic{} shapes
training, not inference. \emph{(iii) Nearly free:} the teacher is the
student's own detached sub-configuration: one extra partial forward on
at most $K$ supervised tokens per micro-step, no external models, no
transfer set, no separate distillation phase (contrast RegMean's
merge-time solves \citep{jin2023regmean}, FuseLLM's continual
distillation corpus \citep{wan2024fusellm}, and post-hoc multi-teacher
repair \citep{mtkd2025}).
Conceptually, \fic{} is the functional projection of the orthogonality
desideratum: rather than demanding $\Delta W_L$ live in a subspace
where it \emph{cannot} affect domain computation anywhere, it demands
$\Delta W_L$ \emph{happen not to} affect domain behavior on the domain
distribution, seemingly the weaker, better-targeted condition. Our
results show the opposite: the ostensibly over-strong subspace
condition is what protects language fidelity and the base model's
competences, while the distribution-limited functional condition
leaves both exposed.

\subsection{Composition at Deployment}
\label{sec:method:merge}

Default deployment is uniform addition
$W = W_0 + \Delta W_D + \Delta W_{L_\ell}$ with no tuned coefficients
(\emph{route-free CAT} in our runs); a module-wise asymmetric
variant motivated by the upper-layer concentration of language
identity \citep{bandarkar2025layer, hinck2024ifl} was cut with the
other reduced-scope items (\S\ref{sec:setup}) and is future work.

\section{Experimental Setup}
\label{sec:setup}

\paragraph{Models, data, budget.}
Base: LLaVA-1.5-7B. \dlora{}: rank 64, on 100k stratified
GeoChat-Instruct \citep{kuckreja2024geochat}. \llora{}$_\ell$: rank
32, on 67k Bactrian-X \citep{li2023bactrianx} for
$\ell \in$ \{es, ar, hi, vi, sw\} (four scripts; Vietnamese and Swahili
low-resource).
\textbf{Compute-bounded protocol:} every trained system, including
all baselines, receives an identical matched budget of 1{,}000
optimizer steps (a preregistered fallback from 2{,}500, executed
before any run passed step 100 when measured A100 throughput made the
larger budget infeasible in the reserved window) with identical data
order, token budget, and stopping rule; core comparisons use 3 seeds;
hyperparameters were frozen from a preregistered es/vi pilot before
any test-set contact. We report GPU-hours for every system. Claims are
explicitly scoped to this budget; the full-epoch regime is future
work.

\begin{table*}[t]
\centering\small
\setlength{\tabcolsep}{3.4pt}
\begin{tabular*}{\textwidth}{@{\extracolsep{\fill}}l ccc ccc ccc cc@{}}
\toprule
& \multicolumn{3}{c}{en} & \multicolumn{3}{c}{es} & \multicolumn{3}{c}{vi} & \multicolumn{2}{c}{Belebele ret.} \\
\cmidrule(lr){2-4}\cmidrule(lr){5-7}\cmidrule(lr){8-10}\cmidrule(l){11-12}
System & Acc$\uparrow$ & LF$\uparrow$ & A$\cap$L$\uparrow$ & Acc$\uparrow$ & LF$\uparrow$ & A$\cap$L$\uparrow$ & Acc$\uparrow$ & LF$\uparrow$ & A$\cap$L$\uparrow$ & es$\uparrow$ & vi$\uparrow$ \\
\midrule
Base LLaVA-1.5 (zero-shot)  & 63.23 & 89.29 & 55.40 & 48.97 & 73.35 & 32.35 & 39.36 & 96.13 & 35.76 & 70.00 & 58.11 \\
Qwen2.5-VL-7B (reference)   & 72.80 & 90.62 & 64.92 & 71.94 & 73.30 & 53.58 & 64.42 & 97.77 & 62.60 & 87.67 & 86.78 \\
GeoChat-7B (RS expert, ref.) & 80.41 & 90.07 & 71.89 & 66.74 & 74.94 & 49.07 & 29.93 & 97.77 & 28.66 & 65.67 & 54.00 \\
\midrule
Best merger: layer swap$^\dagger$ & 65.10 & 87.70 & 58.02 & 45.65 & 50.80 & 13.58 & 42.41 & 75.67 & 27.24 & 46.00 & 33.11 \\
Prior orth.: OSRM$^\dagger$ & 51.69 & 81.39 & 51.62 & 47.06 & 23.92 & 0.00 & 51.62 & 0.32 & 0.05 & 43.78 & 31.00 \\
\textsc{Joint} (J0)         & 50.46 & 88.96 & 45.06 & 47.55 & 32.70 & 13.88 & 32.85 & 45.39 & 7.90 & 29.41 & 27.15 \\
\textsc{Joint} 2$\times$ (2k steps)& 60.73 & 80.41 & 56.54 & 63.33 & 15.95 & 4.33 & 61.82 & 0.00 & 0.00 & 23.78 & 24.78 \\
\fic{} (functional)         & 50.25 & 85.27 & 45.88 & 42.55 & 22.37 & 5.16 & 48.70 & 11.22 & 3.51 & 49.70 & 38.81 \\
\modl{} (ours)        & \textbf{77.57} & \textbf{90.45} & \textbf{68.93} & \textbf{78.66} & \textbf{75.25} & \textbf{58.21} & \textbf{63.45} & \textbf{95.90} & \textbf{59.42} & \textbf{67.93} & \textbf{56.19} \\
\bottomrule
\end{tabular*}
\caption{\textbf{Main results on the GeoChat-Bench classification holdout
(2{,}195 items/language) and Belebele likelihood retention.}
\textsc{Joint}, \fic{}, and \modl{} are means over three seeds (J0
seed 44 diverged; its last healthy checkpoint, step 500, is
substituted and the diverged run reported in App.~\ref{app:full});
the merge, OSRM, and \textsc{Joint} 2$\times$ rows are single runs, as their
scores sit at a floor no seed resample plausibly escapes. Acc = class-identity accuracy (any accepted
surface form); LF = response in the query language; A$\cap$L =
correct and in the query language.
\modl{} is the only matched-budget system that beats the untrained
base on A$\cap$L, and it exceeds the Qwen2.5-VL reference on es. The
\modl{}$-$J0 A$\cap$L deltas are $+23.87$/$+44.33$/$+51.53$
(en/es/vi), item-paired bootstrap over pooled seeds (n=6{,}585, 95\% confidence
intervals $[22.8,25.0]$/$[42.9,45.8]$/$[50.3,52.8]$, all excluding zero). GeoChat-7B (evaluated via a
weight conversion to the llava-hf layout at its native 504\,px) is a
reference, not a matched baseline: it trains on the full 318k
GeoChat-Instruct corpus, roughly $20\times$ our budget.
$^\dagger$Cross-lingual layer swap is the best of 14 training-free
mergers by es/vi A$\cap$L (all 14 in Table~\ref{tab:mergefull}; by the en column alone TIES is
strongest at 68.5: merging preserves English, not the target
language); merge experts predate the EOS fix, so
their generations are scored after repetition collapse under the
disclosed-asymmetry rule fixed in advance. O-LoRA shows the same
failure signature as OSRM---A$\cap$L at the floor on both target
languages (Table~\ref{tab:orthfamily}).}
\label{tab:main}
\end{table*}

\paragraph{Benchmarks and translation.}
RS: GeoChat-Bench scene-classification holdout
\citep{kuckreja2024geochat} (2{,}195 items/language: UCMerced 2{,}100
+ AID 95, main table; VRSBench \citep{li2024vrsbench} is excluded
because its licensed imagery was unavailable, and the region and
hrben/lrben subsets ship no usable references; both exclusions were
preregistered), GEOBench-VLM multiple-choice (MCQ) subset \citep{danish2025geobench}
(3{,}001, external validation).
Language retention: Belebele \citep{bandarkar2024belebele} (900 items
$\times$ 6 languages) and MMLU-ProX Lite \citep{xuan2025mmluprox}
(588 $\times$ 6).
RS benchmarks are translated with NLLB-200-3.3B \citep{costa2022nllb}
under placeholder protection for numbers, coordinates, units, option
letters, and proper names; every translated manifest passes a
structural answer-key integrity audit (choice-label preservation,
record/ID/option alignment, reference--prompt consistency;
App.~\ref{app:audit}). Translation
artifacts and their known effects \citep{park2024translation} are why
we frame this as protocol, not resource.

\paragraph{Metrics.}
\textbf{Acc} is \emph{class-identity} accuracy: the prediction names
the correct class in any accepted surface form (English canonical,
translated reference, or the prompt option list's rendering). This
definition is forced by two defects we quantify: scoring against the
translated reference alone conflates accuracy with answer language,
and our translation pipeline rendered references and prompt option
lists independently, leaving 41.1\%/24.5\% of es/vi references absent
from the offered options. \textbf{LF}, language fidelity (response in the query language,
per language identification, LID); \textbf{Acc$\cap$LF} (correct \emph{and} in-language; the headline
metric, written A$\cap$L in tables); text-language \textbf{retention} via
likelihood-ranked MCQ accuracy on Belebele/MMLU-ProX, which reads one
forward pass and is immune to the generation-side EOS defect described
below. Per-system GPU-hours are in App.~\ref{app:hyper}.
\textbf{EOS control.} Our training initially omitted the EOS
terminator from supervision targets, deflating free-form metrics of
fine-tuned models by up to $7\times$; all main results use
EOS-corrected retraining, likelihood-ranked metrics are immune, and
the defective runs are retained as an item-paired control
(App.~\ref{app:judge}).
Uncertainty is reported as item-paired bootstrap intervals (10k
resamples) for the headline comparison and seed min--max ranges
elsewhere; per-language results are always shown, never only macro
averages.

\paragraph{Systems.}
\emph{References (inference only):} base LLaVA-1.5, official GeoChat
checkpoint, Qwen2.5-VL-7B \citep{wang2024qwen2vl} (incidental
multilinguality bound); closed models are cited, not run.
\emph{Independent-training mergers} (all from the same pair of
independently trained domain/language experts, D-IND/L-IND): Linear, TIES \citep{yadav2023ties},
DARE($+$Linear/$+$TIES) \citep{yu2024dare}, KnOTS-DARE-TIES
\citep{stoica2025knots}, CAT \citep{prabhakar2024cat}, and
cross-lingual layer swap \citep{bandarkar2025layer}; KnOTS-TIES was
excluded when its SVD alignment exceeded a 7-hour wall-clock
budget.\footnote{DO-Merging \citep{zheng2025domerging}
is discussed but excluded from the executable matrix: no verified
official implementation was available, and we do not benchmark against
our own approximations of others' methods.}
\emph{Merge-aware training:} \textsc{Joint} (J0), a compute-matched
\textsc{Joint} at double training steps (\textsc{Joint} 2$\times$), \modl{}
(\S\ref{sec:method:geo}), and \fic{}.
\emph{Prior orthogonal training:} OSRM \citep{zhang2025osrm} and
O-LoRA \citep{wang2023olora} at the same budget. The originally
planned \textsc{Orth-W} (whitened), module-wise composition,
SOS-LoRA, a translate-test pivot, D-only/L-only decompositions, an
Aya data ablation, and the MT-trained upper bound (PALO recipe,
\citealp{maaz2025palo}) were cut when the compute window contracted;
each cut is disclosed here rather than silently dropped, and the
associated claims are removed.

\section{Results}
\label{sec:results}

\begin{table}[t]
\centering\scriptsize
\setlength{\tabcolsep}{3.6pt}
\begin{tabular*}{\columnwidth}{@{\extracolsep{\fill}}l l cccc@{}}
\toprule
Method & Constraint & Acc$\uparrow$ & LF$\uparrow$ & A$\cap$L$\uparrow$ & Bel.$\uparrow$ \\
\midrule
\textsc{Joint}      & none                    & 40.20 & 39.05 & 10.89 & 28.28 \\
OSRM$^\dagger$      & init-time               & 49.34 & 12.12 & 0.03  & 37.39 \\
O-LoRA$^\dagger$    & sequential, one-sided   & 45.40 & 12.89 & 0.03  & 39.84 \\
\modl{} (ours)& mutual, training-long   & \textbf{71.06} & \textbf{85.58} & \textbf{58.82} & \textbf{62.06} \\
\bottomrule
\end{tabular*}
\caption{\textbf{Which orthogonality?} The orthogonal-training family under
one matched budget, es/vi average (holdout Acc/LF/A$\cap$L and
Belebele retention, \%). $^\dagger$Built on the pre-fix domain expert and scored
with repetition collapse (App.~\ref{app:judge}); their
likelihood-based Belebele columns are defect-immune.}
\label{tab:orthfamily}
\end{table}

\paragraph{Which orthogonality matters.}
Table~\ref{tab:orthfamily} decomposes the recipe's two design
choices. OSRM separates the subspaces once, at initialization, and
training freely re-entangles them; O-LoRA constrains only the
language adapter against a frozen domain adapter. Both raise Belebele
retention over \textsc{Joint} (the constraint does protect text
competence) yet leave A$\cap$L at zero: fidelity is lost unless
\emph{both} updates are kept apart \emph{while they are learned}.
The mutual, training-long constraint is therefore not an
implementation detail but the operative ingredient.

\begin{figure}[t]
\centering
\includegraphics[width=\columnwidth]{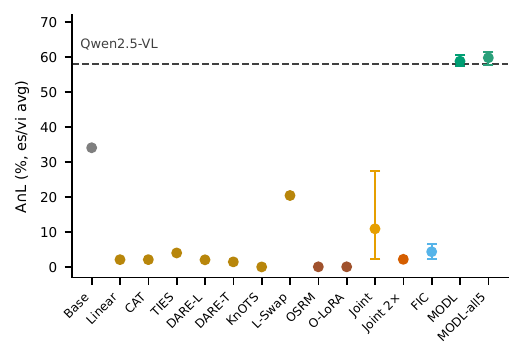}
\caption{\textbf{All systems on one axis (A$\cap$L, es/vi average).}
Markers are three-seed means at the primary hyperparameter; error
bars span the seed minimum--maximum (systems trained once have no
bar; $\lambda$ ablations are in Table~\ref{tab:ablation}). Both \modl{} variants sit at the Qwen2.5-VL reference line with
seed ranges of a few points; every alternative lies at or below the
untrained base. \textsc{Joint}'s wide bar reflects its diverged seed's
step-500 substitute.
The full per-language, per-seed breakdown is in
Table~\ref{tab:perseed} (App.~\ref{app:full}).}
\label{fig:seeds}
\end{figure}

\begin{figure*}[t]\centering
\includegraphics[width=\textwidth]{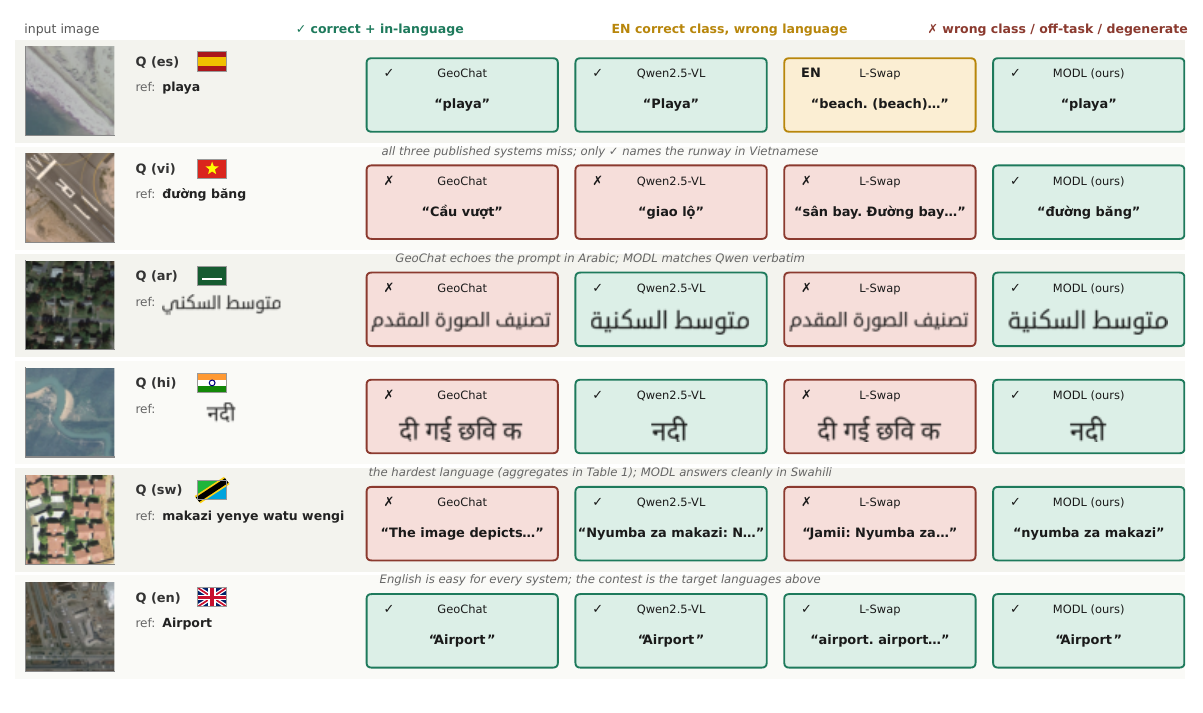}
\caption{\textbf{Case studies across all five languages against
published systems:} GeoChat-7B \citep{kuckreja2024geochat}, Qwen2.5-VL-7B
\citep{wang2024qwen2vl}, and the cross-lingual layer-swap merger
\citep{bandarkar2025layer}. All outputs are verbatim generations on true satellite
inputs. Box colors: green = correct class in the query language;
orange = correct class in English; red = wrong class, off-task, or
degenerate. On vi all three published systems miss while \modl{}
answers correctly; on ar/hi GeoChat echoes the prompt and \modl{}
matches Qwen's in-language answer verbatim with zero
multilingual--multimodal data; on Swahili, the hardest language in
aggregate (Table~\ref{tab:main}), \modl{} answers cleanly in
Swahili while GeoChat drifts to an English description. The final
row shows English, where every system succeeds: the contest is
entirely in the target languages. Cases illustrate individual failure
modes; aggregate comparisons are in Table~\ref{tab:main}.}
\label{fig:cases}
\end{figure*}

\paragraph{Preregistered decision rules and outcomes.}
Figure~\ref{fig:seeds} places every trained system on one axis; the
seed-level spread behind each mean below can be read from it directly,
and the full per-language breakdown is in App.~\ref{app:full}. Three
falsifiable predictions were frozen before the confirmatory runs (the
registration originally favored the functional hypothesis;
App.~\ref{app:hyper}), and the data answered all three.
\emph{(i) Geometry.} Following text-only evidence
\citep{zhang2025rethinking}, the registration predicted that mutual
orthogonality would not beat \textsc{Joint}. The data show the
opposite: \modl{} improved A$\cap$L to $0.56$--$0.71$ against
${\le}0.05$ for \textsc{Joint} on es/vi on both stable seeds (the
diverged seed's step-500 substitute peaks at $0.33$, still ${\ge}24$
points below \modl{}), with paired bootstrap intervals excluding
zero.
\emph{(ii) Function space.} The registration predicted \fic{} would
beat \textsc{Joint} and the geometric variants. \fic{} improved
Belebele/MMLU retention over \textsc{Joint} in all six cells on all
three seeds, but recovered neither language fidelity nor
target-language accuracy, and \modl{} exceeded it on every axis.
\emph{(iii) Language spread.} The registration predicted the largest
gains for vi/sw. Instead, the five-language variant preserved
en/es/vi (A$\cap$L $0.68$/$0.60$/$0.56$) and collapsed ar/hi/sw.
Per-pair runs rule out budget dilution as the cause: trained on one
language each, \modl{} matches the two-language joint deployment on
es/vi (A$\cap$L $59.2$/$55.7$), restores full in-language answering
for ar/hi (LF $1.0$ vs.\ $0.48$/$0.15$ for per-pair
\textsc{Joint}), lifts their A$\cap$L to $15.9$/$23.6$ (vs.\
$0.2$/$3.4$), and returns Belebele retention to base level, yet
in-language class naming stays far below the es/vi $56$--$59$. The constraint's repairs are
universal; what remains is a script and base-coverage boundary on
naming RS classes in non-Latin scripts, with Swahili at the floor for
every system.
The registered advance criterion against the best merger could not be
evaluated cleanly (the merge experts predate the EOS fix; a clean
retrain did not fit the compute window) and is reported under the
disclosed-asymmetry rule fixed in advance (App.~\ref{app:full}).
One unplanned observation: under the shared no-clipping recipe,
unconstrained training diverged in 3 of 7 runs while all
orthogonality-constrained runs were stable; we report this
descriptively.

\paragraph{Ablations.}
Table~\ref{tab:ablation} gathers all ablations with an explicit
grouping column: constraint strength, the \fic{} consistency weight
(no value rescues fidelity or accuracy), the unconstrained controls,
and the language-scope question of the same recipe deployed five-way
versus per-pair. The originally planned wider grid
(whitened penalty, module-wise composition, projector access, rank
grid, +Aya) was cut under the compute window and is listed in
Limitations rather than silently dropped.

\begin{table}[t]
\centering\scriptsize
\setlength{\tabcolsep}{2.5pt}
\begin{tabular*}{\columnwidth}{@{\extracolsep{\fill}}l l l ccc@{}}
\toprule
Ablation & Variant & Lang. & Acc$\uparrow$ & A$\cap$L$\uparrow$ & Bel.$\uparrow$ \\
\midrule
\multirow{3}{*}{\shortstack[l]{ours: orthogonality\\knob $\lambda_{\perp}$}}
 & 0.1 (main) & es+vi & 69.59 & 57.47 & 62.11 \\
 & 0.01       & es+vi & 68.50 & 56.40 & 58.06 \\
 & 1.0        & es+vi & 66.99 & 54.65 & 64.06 \\
\midrule
\multirow{3}{*}{\shortstack[l]{\fic{}: consistency\\knob $\lambda_c$}}
 & 0.5        & es+vi & 51.00 & 4.28  & 43.50 \\
 & 0.25       & es+vi & 36.15 & 3.67  & 49.22 \\
 & 1.0        & es+vi & 51.12 & 6.01  & 46.94 \\
\midrule
\multirow{2}{*}{\shortstack[l]{no constraint\\($\lambda_{\perp}{=}\lambda_c{=}0$)}}
 & \textsc{Joint}    & es+vi & 43.46 & 2.16 & 32.44 \\
 & \textsc{Joint} 2$\times$ steps & es+vi & 62.57 & 2.16 & 24.28 \\
\midrule
\multirow{5}{*}{scope: five-way}
 & \multirow{5}{*}{one adapter} & es & 80.18 & 59.95 & 67.89 \\
 & & vi & 59.54 & 55.58 & 56.56 \\
 & & ar & 6.38  & 6.38  & 41.33 \\
 & & hi & 12.85 & 12.85 & 35.22 \\
 & & sw & 3.05  & 0.00  & 33.11 \\
\midrule
\multirow{5}{*}{scope: per-pair}
 & \multirow{5}{*}{one lang.\ each} & es & 79.77 & 59.23 & 67.89 \\
 & & vi & 59.77 & 55.72 & 55.89 \\
 & & ar & 15.85 & 15.85 & 43.56 \\
 & & hi & 23.55 & 23.55 & 36.44 \\
 & & sw & 20.87 & 3.01  & 34.11 \\
\bottomrule
\end{tabular*}
\caption{\textbf{Unified ablations (seed 43).} Rows above the scope blocks are
es/vi averages; scope rows are per-language cells of the same recipe
deployed five-way (one shared adapter) or per-pair (one language per
adapter). Latin-script languages survive even five-way sharing;
per-pair training doubles ar/hi; doubling \textsc{Joint}'s budget
raises Acc but leaves A$\cap$L at floor: compute does not substitute
for the constraint.}
\label{tab:ablation}
\end{table}

\section{Analysis}
\label{sec:analysis}

\paragraph{Geometry vs.\ function.}
The two constraints dissociate cleanly: \modl{} separates the subspaces and is the
only system whose composition works, while \fic{} leaves geometry
essentially unchanged and fails on fidelity and accuracy. In this
multimodal setting, subspace separation empirically \emph{is} the
operative mechanism, the opposite of the text-only picture drawn by
\citet{zhang2025rethinking}. Consistent with this, \modl{} is
also the only matched-budget system whose text-side likelihoods stay
at the untrained base's level, suggesting the constraint protects the base's
representations wholesale rather than trading capabilities.

\paragraph{Interference matrix.}
Table~\ref{tab:main} doubles as a capability-retention matrix, with
English RS, target-language RS, and target-language text as its axes:
naive training and every prior composition route sacrifice at least
one axis, and \modl{} is the only intervention positive on all
three.

\paragraph{Where do domain and language live?}
The merger ranking offers indirect evidence that the text-only
picture, language identity concentrating in upper layers
\citep{bandarkar2025layer}, transfers to MLLMs: the only merger with
non-trivial target-language fidelity is the one that swaps upper
layers wholesale (layer swap, A$\cap$L 27.2 on vi), while every
weight-mixing merger is at the floor
(Table~\ref{tab:mergefull}). A systematic layer-band sweep is left
to future work.

\paragraph{Case studies.}
Figure~\ref{fig:cases} walks the five languages on real holdout
items against the published systems: per-pair \modl{} answers
correctly in Spanish, Vietnamese, Arabic, and Hindi, Swahili sits at
the script boundary, and the failure modes match the aggregates
(prompt echo, English drift, off-task text).

\section{Conclusion}
A remote-sensing MLLM can acquire a new language without a single
multilingual RS example, but only if interference is controlled at
the source. Naive joint training of a domain and a language LoRA
answers correctly in the wrong language, erases text-only
multilingual competence, and is unstable across seeds; a
function-space variant softens only the second failure, and published
orthogonal-training methods fail outright. One symmetric
mutual-orthogonality loss term resolves all three on every seed,
surpassing the untrained base and, on Spanish scene classification,
the Qwen2.5-VL reference: in this multimodal setting, parameter
geometry is the deciding factor. The recipe applies wherever an
English-only domain MLLM and target-language text instructions
exist; code, adapters, and the preregistered protocol are
released.

\section*{Limitations}
Claims are scoped to a matched compute-bounded budget (1{,}000 steps
per system, 6.8\% of the assembled corpus; the preregistered 2{,}500
and the plan's larger ambitions did not fit the reserved window);
full-convergence behavior may differ. The domain evaluation is one
benchmark family (scene classification); harder open-ended RS tasks
may behave differently. The independent-merge baselines are built from
experts trained before the EOS fix and are reported under a
disclosed-asymmetry rule fixed in advance, as are OSRM and O-LoRA,
whose frozen domain expert predates the fix; SOS-LoRA, the MT-trained
upper bound, the translate-test pivot, and the whitened-orthogonality
and module-wise variants were cut and not run. The shared recipe uses no gradient
clipping; the observed stability advantage of the orthogonality
constraint is reported under that recipe and may shrink with clipping.
The five-language joint variant dilutes low-resource languages at this
budget, so the recipe claim is per language pair. Language fidelity is
judged by an n-gram LID over short class-name answers; its error on
such fragments is not separately validated. All non-English numbers
rest on NLLB translation quality; no native-speaker validation was
performed.
Evaluation uses machine translation, not natively authored RS
test sets, which do not exist beyond EN/ZH; artifacts cannot be fully
excluded \citep{park2024translation}. The recipe presumes an
English-centric base; on multilingual backbones the language module
may be partially redundant (our Qwen2.5-VL reference bounds this).
Text-only language LoRAs teach language form, not culturally grounded
visual concepts. We study one domain, five languages, one 7B
architecture; the exploratory-to-confirmatory protocol and
preregistered gates are reported so that negative or null confirmatory
outcomes are interpretable rather than hidden.

\section*{Ethics Statement}
Cross-lingual RS assistants lower barriers for non-English-speaking
communities in disaster response and land management; the same
capability could ease misuse in more languages, though no imaging
capability beyond the English base is added.
Machine-translated evaluation data inherits NLLB-200 biases. All datasets and models used are public
research releases; licenses are respected; no paid closed-model APIs
are used anywhere in the experimental pipeline.

\bibliographystyle{acl_natbib}
\bibliography{refs}

\appendix

\section{Translation Protocol}
\label{app:audit}
All evaluation sets are translated with a pinned local NLLB-200-3.3B
under placeholder protection for numbers, coordinates, units, option
letters, and proper names, with SHA-256-checked resumable batching; no
paid APIs are used anywhere. Two pipeline defects were discovered and
quantified during the confirmatory phase (\S\ref{sec:setup}): the
class inventory was rendered independently for references and for
prompt option lists, leaving 41.1\%/24.5\% of es/vi references absent
from the offered options; scoring therefore uses the released
class-identity map (option-list-aligned accepted surface forms,
2{,}195/2{,}195 positional alignment in all five languages).

\section{Hyperparameters, Compute, and Preregistration}
\label{app:hyper}
LLaVA-1.5-7B, 4-bit NF4 base, bf16 compute; \dlora{} rank 64 on LLM
attention/MLP and projector, \llora{} rank 32 on the LLM only; AdamW,
lr $2{\times}10^{-4}$, no gradient clipping, micro-batch 1,
gradient accumulation 16, max length 1{,}024, 1{,}000 optimizer steps
(preregistered fallback from 2{,}500, executed before any run passed
step 100), checkpoints every 250, seeds 43/44/45. \fic{}:
$\lambda_c{=}0.5$ primary ($0.25/1.0$ ablated), $K{=}64$, $\tau{=}1$.
\modl{}: $\lambda{=}0.1$ primary ($0.01/1.0$ ablated), factor
cosine, sum reduction. Approximate cost: J0 $\approx$7 GPU-h/run, \fic{}
$\approx$14 (detached-teacher forwards), \modl{} $\approx$7.3.

\paragraph{Compute.} All experiments ran on NVIDIA A100-SXM4-40\,GB
GPUs (a Slurm partition of three 4-GPU nodes), one GPU per job; no
multi-GPU training was used. Slurm accounting for the complete study
records $\approx$1{,}249 A100-hours: $\approx$501 for the final
reported trainings (the matched-budget systems across seeds, plus the
independent domain/language experts behind the mergers),
$\approx$325 for evaluation, $\approx$109 for the $\lambda$
ablations and the cut SOS-LoRA pilot, $\approx$115 for the
superseded no-EOS trainings retained as controls, and $\approx$199
for failed, cancelled, or timed-out jobs (including merge-construction
jobs that exceeded their wall-clock).

\paragraph{Implementation.} Python 3.12.13, PyTorch 2.2.2+cu121,
Transformers 4.57.3, PEFT 0.19.1 (LoRA training and composition),
bitsandbytes 0.43.3 (NF4 quantization). Translation uses the pinned
local \texttt{facebook/nllb-200-3.3B} checkpoint with greedy
decoding (\texttt{do\_sample=False}, no beam search, forced
target-language BOS, source truncation and \texttt{max\_new\_tokens}
at 512, batch size 8). Language identification uses langid 1.1.6
constrained to the six study languages. Evaluation-time generation is
greedy with at most 16 new tokens; likelihood metrics use a single
forward pass and no sampling.

The preregistration (frozen gates, 22 addenda covering the
environment incident, the budget fallback, the EOS defect and
retraining, divergence handling, the merge-asymmetry protocol, and
this closeout) is released with the code. The paper was originally
registered around \fic{}; the reversal to \modl{} is documented
in Addenda 19--22, each written before the corresponding results
existed.

\paragraph{Input--output formats.}
\emph{Domain stream} (updates \dlora{}): input is one RS image with an
English instruction in the LLaVA chat template,
\texttt{<image> USER: Classify the given image in one of the
following classes. Classes: \dots\ ASSISTANT:}; the supervised target
is the English answer followed by the EOS token
(\texttt{tennis court</s>}), with loss on the answer span only.
\emph{Language stream} (updates \llora{}$_\ell$): input is a
text-only target-language instruction from Bactrian-X, no image; the
target is the target-language response plus EOS. The two streams
alternate mini-batches; no example ever pairs a target language with
an image. \emph{Evaluation}: holdout inputs are an RS image plus the
translated classification prompt (e.g.\ \texttt{<image> USER:
Clasifique la imagen dada \dots\ ASSISTANT:}); generation is greedy
with at most 16 new tokens. Belebele/MMLU-ProX inputs are text-only
MCQs; no generation is used there---the answer is the option letter
with the highest next-token likelihood in a single forward pass.

\section{Judging, LID, and the EOS Control}
\label{app:judge}
Scoring is fully deterministic: normalized exact match and token-F1,
option-letter extraction for MCQ, likelihood-ranked MCQ from a single
forward pass, and class-identity accuracy against the accepted-form
map. Language identification uses langid constrained to the six study
languages; answers ineligible for LID (numbers, option letters) are
reported as coverage. For systems trained before the EOS fix
(mergers, the frozen domain expert inside OSRM/O-LoRA), generations
are first collapsed to their initial non-repeating segment, a rule
fixed in Addendum 21 before any such result existed. The EOS control
itself: training without a supervised terminator leaves free-form
generation running to the token limit ($35$--$41\%$ of items at
1{,}000 steps), deflating token-F1 by up to $7\times$
(en 8.5$\to$60.5 for J0, same seed and data), while
single-forward-pass likelihood metrics are unaffected; the archived
no-EOS twin of every run permits item-paired comparison.

\section{Full Results}
\label{app:full}

Per-seed holdout results (Acc/LF/A$\cap$L, \%):

\begin{table*}[p]\centering\small
\setlength{\tabcolsep}{4.5pt}
\begin{tabular*}{\textwidth}{@{\extracolsep{\fill}}l ccc ccc ccc@{}}
\toprule
& \multicolumn{3}{c}{en} & \multicolumn{3}{c}{es} & \multicolumn{3}{c}{vi} \\
\cmidrule(lr){2-4}\cmidrule(lr){5-7}\cmidrule(l){8-10}
System & Acc & LF & A$\cap$L & Acc & LF & A$\cap$L & Acc & LF & A$\cap$L \\
\midrule
Base LLaVA-1.5 & 63.2 & 89.3 & 55.4 & 49.0 & 73.3 & 32.3 & 39.4 & 96.1 & 35.8 \\
GeoChat-7B & 80.4 & 90.1 & 71.9 & 66.7 & 74.9 & 49.1 & 29.9 & 97.8 & 28.7 \\
Qwen2.5-VL-7B & 72.8 & 90.6 & 64.9 & 71.9 & 73.3 & 53.6 & 64.4 & 97.8 & 62.6 \\
J0 s43 & 54.8 & 90.1 & 50.3 & 58.0 & 11.3 & 4.3 & 28.9 & 31.8 & 0.0 \\
J0 s44 (diverged) & 0.0 & 100.0 & 0.0 & 0.0 & 0.0 & 0.0 & 0.0 & 0.0 & 0.0 \\
J0 s44 @step500 & 43.2 & 88.9 & 35.9 & 43.5 & 64.1 & 32.8 & 31.7 & 66.5 & 22.2 \\
J0 s45 & 53.4 & 87.9 & 48.9 & 41.2 & 22.7 & 4.5 & 37.9 & 37.8 & 1.5 \\
\textsc{Joint} 2$\times$ s43 & 60.7 & 80.4 & 56.5 & 63.3 & 15.9 & 4.3 & 61.8 & 0.0 & 0.0 \\
\fic{} s43 & 55.7 & 88.4 & 51.1 & 48.3 & 20.8 & 5.8 & 53.7 & 5.8 & 2.8 \\
\fic{} s44 & 47.6 & 83.4 & 43.4 & 38.5 & 29.8 & 5.3 & 46.8 & 26.2 & 7.7 \\
\fic{} s45 & 47.5 & 84.0 & 43.1 & 40.9 & 16.4 & 4.4 & 45.6 & 1.7 & 0.0 \\
\fic{} $\lambda_c{=}.25$ & 40.3 & 79.9 & 32.5 & 38.7 & 30.3 & 4.3 & 33.6 & 33.2 & 3.0 \\
\fic{} $\lambda_c{=}1$ & 57.6 & 87.7 & 51.4 & 50.7 & 28.3 & 4.9 & 51.5 & 29.2 & 7.2 \\
\modl{} s43 & 76.4 & 90.5 & 67.9 & 79.0 & 73.9 & 58.8 & 60.1 & 95.9 & 56.1 \\
\modl{} s44 & 79.8 & 89.9 & 70.9 & 78.9 & 75.5 & 57.8 & 67.1 & 96.0 & 63.1 \\
\modl{} s45 & 76.5 & 90.9 & 68.0 & 78.0 & 76.3 & 58.0 & 63.1 & 95.8 & 59.0 \\
\modl{} $\lambda{=}.01$ & 72.3 & 90.5 & 64.1 & 73.9 & 73.3 & 54.0 & 63.1 & 94.9 & 58.8 \\
\modl{} $\lambda{=}1$ & 80.2 & 91.0 & 71.8 & 75.4 & 75.9 & 55.2 & 58.5 & 95.3 & 54.1 \\
\bottomrule
\end{tabular*}
\caption{\textbf{Per-seed GeoChat-holdout results.} The diverged J0 s44 is retained as the
literal record; step-500 is its substituted baseline.}
\label{tab:perseed}
\end{table*}

Five-language \modl{} across seeds (A$\cap$L, \%):

\begin{table}[h]\centering\scriptsize
\begin{tabular*}{\columnwidth}{@{\extracolsep{\fill}}l cccccc@{}}
\toprule
 & en & es & ar & hi & vi & sw \\
\midrule
\modl{}-all5 s43 & 68.4 & 60.0 & 6.4 & 12.8 & 55.6 & 0.0 \\
\modl{}-all5 s44 & 69.8 & 57.5 & 11.8 & 17.3 & 62.6 & 0.0 \\
\modl{}-all5 s45 & 70.7 & 59.5 & 9.6 & 11.0 & 63.6 & 0.0 \\
\bottomrule
\end{tabular*}
\caption{\textbf{Five-way joint training preserves en/es/vi on all
three seeds and dilutes ar/hi/sw.} ar/hi answer in-language (LF ${\ge}99$)
but misname the class.}
\label{tab:all5seeds}
\end{table}

All fourteen mergers (collapsed scoring; Acc/A$\cap$L per cell, \%):

\begin{table*}[p]\centering\small
\setlength{\tabcolsep}{6pt}
\begin{tabular*}{\textwidth}{@{\extracolsep{\fill}}l cc cc cc cc@{}}
\toprule
& \multicolumn{2}{c}{es$\to$en} & \multicolumn{2}{c}{es$\to$es} & \multicolumn{2}{c}{vi$\to$en} & \multicolumn{2}{c}{vi$\to$vi} \\
\cmidrule(lr){2-3}\cmidrule(lr){4-5}\cmidrule(lr){6-7}\cmidrule(l){8-9}
Merger & Acc & A$\cap$L & Acc & A$\cap$L & Acc & A$\cap$L & Acc & A$\cap$L \\
\midrule
cat & 63.0 & 58.7 & 52.5 & 4.1 & 61.5 & 57.4 & 57.0 & 0.0 \\
linear & 63.0 & 58.7 & 52.5 & 4.1 & 61.5 & 57.4 & 57.0 & 0.0 \\
ties & 75.3 & 68.0 & 52.4 & 4.6 & 75.1 & 68.5 & 60.1 & 3.3 \\
dare\_linear & 53.2 & 50.3 & 39.0 & 4.1 & 55.5 & 51.2 & 48.6 & 0.0 \\
dare\_ties & 54.5 & 50.6 & 31.1 & 2.9 & 60.8 & 56.5 & 47.4 & 0.0 \\
knots\_dare\_ties & 10.7 & 10.7 & 8.8 & 0.2 & 13.6 & 13.5 & 2.2 & 0.0 \\
layer\_swap & 65.0 & 57.9 & 45.6 & 13.6 & 65.2 & 58.1 & 42.4 & 27.2 \\
knots\_ties & \multicolumn{8}{c}{merge exceeded 7h wall-clock (SVD); excluded, disclosed} \\
\bottomrule
\end{tabular*}
\caption{\textbf{Every merger transfers capability (en columns) and
fails language fidelity (target-language A$\cap$L), mirroring naive
joint training.} OSRM/O-LoRA share this signature: A$\cap$L ${\le}0.05$ on both
target languages (Table~\ref{tab:orthfamily}).}
\label{tab:mergefull}
\end{table*}

\begin{figure}[t]
\centering
\includegraphics[width=\columnwidth]{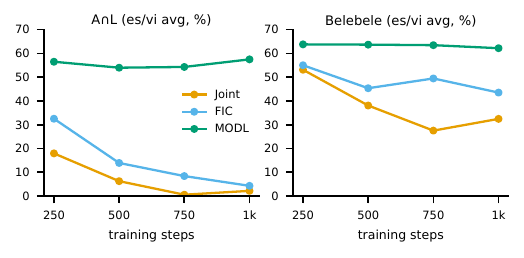}
\caption{\textbf{Checkpoint curves (seed 43, es/vi average).} \textsc{Joint}
decays monotonically on both axes as training proceeds
(A$\cap$L $17.9\to0.6$, Belebele $53\to28$); \fic{} slows but does
not stop the decay; \modl{} is at full strength from step 250 and
flat thereafter. Interference accumulates with optimization, and the
mutual constraint changes the trajectory, not just the endpoint;
this also rules out under-training as an explanation for the
1{,}000-step results.}
\label{fig:stepcurve}
\end{figure}

Figure~\ref{fig:stepcurve} traces all three systems across
checkpoints: naive fidelity destruction accumulates with training
while \modl{} is flat from one quarter of the budget, which is why
budget-matched comparisons are essential and why the 1{,}000-step
scope is not an under-training artifact.

\section{Case Studies}
\label{app:cases}
Figure~\ref{fig:cases} shows representative items with the true
satellite inputs.

\end{document}